\documentclass[twocolumn]{article}
\usepackage{ijcai20}

\usepackage{times}
\usepackage{soul}
\usepackage{url}
\usepackage[hidelinks]{hyperref}
\usepackage[utf8]{inputenc}
\usepackage[small]{caption}
\usepackage{graphicx}
\usepackage{bm}
\usepackage{amsmath}
\usepackage{amssymb}
\usepackage{amsthm}
\usepackage{booktabs}
\usepackage{algorithm}
\usepackage{algorithmic}
\usepackage{multirow}
\title{Scene Text Recognition With Finer Grid Rectification}

\author{
    Gang Wang
    \affiliations
    Beijing Institute of Technology, China
}

\begin{document}

\maketitle

\begin{abstract}
  Scene Text Recognition is a challenging problem because of irregular styles and various distortions. This paper proposed an end-to-end trainable model consists of a finer rectification module and a bidirectional attentional recognition network(Firbarn).
  The rectification module adopts finer grid to rectify the distorted input image and the bidirectional decoder contains only one decoding layer instead of two separated one. Firbarn can be trained in a weak supervised way, only requiring the scene text images and the corresponding word labels. With the flexible rectification and the novel bidirectional decoder, the results of extensive evaluation on the standard benchmarks show Firbarn outperforms previous works, especially on irregular datasets.
\end{abstract}

\section{Introduction}

Scene Text Recognition (STR) is to recognize the word or character sequence in a natural image, which is very primary in many practical applications such as image retrieval and travel navigation. The research of robust scene text recognition remains challenging largely due to the irregular shapes such as perspective and curved text, and distorted patterns of the character.

RARE\cite{shi2016robust} adopts a rectification network before recognition, ASTER\cite{shi2018aster} follows RARE, and ESIR\cite{zhan2019esir} improves the network. The rectification module based on spatial transform network (STN) \cite{jaderberg2015spatial} predicts the text outlines in a weakly supervised way. Ideally, the distorted scene text image is rectified into regular forms. However, ASTER predicts the control points separately and ESIR need to calculate the intersection of predicted line and curve which is time consuming.

\begin{figure}[htbp]
\centering
\includegraphics[width=.4\textwidth]{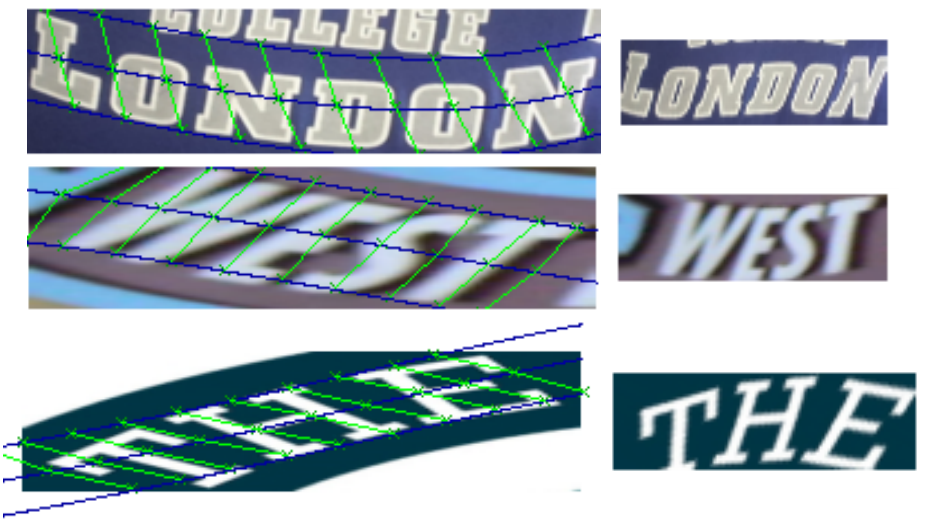}
    \caption{Examples from test data set. Left side is the original image and the blue curve is predicted with STN, then comes the controlpoints. Right side is the rectified image.}
    \label{examples}
\end{figure}

To improve the rectification performance of distorted input image, we proposed a method that takes advantage of finer grid in which the control points are distributed along a curve. Without any character-level or other labeling data, this rectification can generate the control points with smoother boundaries. Figure \ref{examples} shows the result on a grid of size $3 \times 10$.

For bidirectional decoder in STR, ASTER\cite{shi2018aster} have implemented the left-to-right and right-to-left decoding module independently, which means the two decoders are optimized separately. The works of MORAN\cite{cluo2019moran} and ESIR\cite{zhan2019esir} all follow ASTER. However, the forward and backward decoding are (almost) identical and shares much common knowledge, the separate optimization may cause redundancy and low efficiency. Another improvement in this paper is to cut down the two independent forward and backward decoding layer into only one layer network.

The main contributions of this paper is listed as follows: 
1) The finer rectification network is more flexible for scene text distortion modeling and correction.
2) A novel bidirectional decoder is equal or outperforming than the two dependent opposite direction decoder with only one decoding network.
3) With the rectification method and simplified decoder, this end-to-end model achieves state-of-the-art performance on a number of standard benchmarks.

\section{Related Work}

Existing works on STR can be roughly divided into traditional and deep learning based methods.

Most traditional STR work follow a bottom-up pipeline that first detects and recognizes individual character and then links up the recognized characters into words or text lines by language models. For example, \cite{bissacco2013photoocr} uses a fully connected network for character recognition and \cite{jaderberg2014deep} uses CNNs to recognize unconstrained character. These bottom-up methods need to localize each individual characters, which is costly both for location labeling and training. Besides, these methods also prone to errors such as overlaps between adjacent characters.

Deep learning methods have dominated STR in recent years. \cite{jaderberg2016reading} start to take STR as a word classification problem by CNN model which is constrained to the pre-defined vocabulary. Later, various sequence-to-sequence models \cite{shi2016an,shi2016robust,shi2018aster,zhan2019esir}, which is believed to embed the language model in the decoding layer, are applied for STR. Connectionist Temporal Classification (CTC)\cite{shi2016an} in the decoding layer is later changed into attention based models\cite{cheng2018aon,shi2018aster}. VGG\cite{simonyan2014very} in the feature extraction layer is later tended into ResNet\cite{he2016deep}. FAN\cite{cheng2017focusing} developed a focusing attention mechanism to improve the performance of general attention-based encoder-decoder framework.

The rectification in STR models based on the spatial transform network (STN), which improve the recognition accuracy while require no hand crafted features or extra annotations. \cite{risnumawan2014robust} adopts the thin-plate spline (TPS) transformation based on STN for scene text distortion correction. ASTER uses TPS based on STN to rectify the warped image, the points on two sides are predicted without any constraint, two independent decoders was exploited. ESIR extends the rectification to an iterative way, however, the two points on each of the predicted line is not easy to calculate because the root may not exist, which happens when the line and curve doesn't intersect with each other. Besides, the iteration may be not necessary.
 
\section{The Proposed Method}

This section presents the proposed recognition model for scene text including grid rectification module, sequence recognition network and the implementation details.

\subsection{Rectification Network}

thin plate spline based Spatial Transformation Network(STN) is used in our rectification module. The prediction of control points $C \in \mathbb R^{M\times N \times 2}$ is essential to the output of the rectification network. $C=\{C_1, C_2, \cdots, C_M\}, C_i \in \mathbb R^{N \times 2}$.

\subsubsection{Smooth Grid Localization}

Since most characters in 2D scene text images are along a straight line or a smooth curve, the control points has the same trend, a polynomial curve is effective to estimate the tendency of text layout. We can use the curve with a bias to estimate each line of control points $C_i$.

The text line tendency along the characters can be predicted with a polynomial of order $W$ as follow:
\begin{equation}
\begin{split}
    \bar y(x) =&\bm a \bm X = \sum_{k=1}^W{a_k*x^k} \\
    =& a_1 \cdot x+a_2\cdot x^2+\cdots + a_W\cdot x^W
\end{split}
\end{equation}
where $\bm a=(a_1,a_2,\cdots,a_W), \bm X=(x,x^2,\cdots, x^W)$.

With the i-th line of x axis coordinate of the control points and a different bias $b_i$ generated by STN localization network, the j-th y axis coordinate of the control points on i-th line can be determined by: 
\begin{equation}
\begin{split}
    y_j =& b_i + \bar y(x_{ij})= b_i + \bm a \bm X_{ij}\\
    =&b_i + \sum_{k=1}^W{a_k \cdot x_{ij}^k}
\end{split}
\end{equation}
where $\bm a$ is the same between each $C_i$. After this elementwise operation, we can get the whole coordinate of each line of control points $C_i = \{ (x_j, y_j) \}, i \in \{1,2,\cdots, M\}, j \in \{1,2,\cdots, N\}$.

These control points in $C$ divide the 2D plane into $M \times N$ grids, and if the x and y coordinates are predicted directly, the parameters needed is $2M \times N$, however, $M \times N + W$ parameters is enough with the proposed method for $W \ll M \times N$. The tendency of the whole control points stay the same while keeping the text line smooth.

Another intuitive way to see how these points are generated is that they are sampled from the predicted lines, and the lines are parallel to each other for the only difference is the bias.

\begin{table}[htbp]
    \centering
    \begin{tabular}{lll}
        \hline
        Layers  & Out Size & Configurations \\
        \hline
        conv1  & $32 \times 64$ & 32     \\
        conv2  & $16 \times 32$ & 64   \\
        conv3  & $8\times 16$  & 64 \\
        conv4  & $4\times 8$ & 128 \\
        conv5  & $2\times 4$ & 128 \\
        conv6  & $1\times 2$ & 256 \\
        FC1   & 256  &     \\
        FC2   & $ M\times N + M + W $ &      \\
        \hline
    \end{tabular}
    \caption{The structure of the Rectification Network. Each conv layer is followed by a Relu and maxpooling layer, conv's kernel size is $3 \times 3$}
\end{table}

\subsubsection{TPS based Rectification}
The target points $\bm C' = \{\bm C'_i\}, 1\le i \le M$ can be initialized as 
\begin{equation}
\bm C'_i = \left \{ [\frac j N, \frac i M] , 1\le j \le N \right\}
\end{equation}
After we get the control points $\bm C$, the transformation mapping matrix $\bm T\in \mathbb R^{(M\times N +3)\times 2}$ can be calculated by

\begin{equation}
\bm T = \left[ \begin{matrix}
\bm S& \bm 1_{M\times N}& \bm C \\
\bm 1^T_{M\times N}& \bm 0 & \bm 0 \\
\bm C^T & \bm 0& \bm 0 \\
\end{matrix} \right]^{-1} \cdot \left[\begin{array}{c}\bm C' \\\bm 0 \\ \bm 0 \end{array}\right]
\end{equation}
where $\bm S = \{s_{ij}\} \in \mathbb R^{(M\times N) \times (M\times N)}$, $s_{ij} = \phi(\Delta c)$, $\Delta c$ is the difference of every two element in $\bm C$ and $\phi(c) = c^2\log c$. the $\bm C$ here has the size of $(M\times N)\times 2$.

For every pixel $p'$ from the warped image, the corresponding pixel $p$ on the rectified image is
\begin{equation}
\bm p' = \bm T \left[ \begin{array}{c}
1\\
\bm p\\
\phi(\bm p- \bm c_{11})\\
\cdots\\
\phi(\bm p-\bm c_{M\times N})\\
\end{array} \right]
\end{equation}

Although this can be seen as an extension of the two-line control points adopted by Aster and Esir, this method only need to directly predict half amount of coordinates besides the weight and bias parameters. Another advantage of this method is that the generated coordinates maintain a smooth boundary. These two features do help when the rows and columns of the control points increase.

\subsection{Recognition Network}
\begin{figure}[htbp]
    \centering
    \includegraphics[width=.35\textwidth]{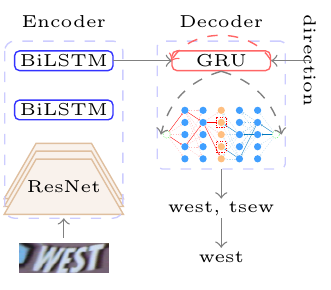}
    \caption{Framework of Recognition Network. Here is a illustration of k=2 beam search for bidirectional decoding. the red arrows denote forward decoder and the blue is backward. the two orange nodes in red dashed boxes is the current candidates.}
    \label{frame}
\end{figure}

The whole pipeline of the recognition network mainly follows the work of Aster\cite{shi2018aster}. Considering the redundant two independent opposite-direction decoder, we simply exploit one decoding layer with a direction embedding to replace the two independent decoder. Figure \ref{frame} is a demonstration of the recognition network architecture.

\subsubsection{Encoder}
The encoder is a common CRNN architecture with ResNet as the feature extractor followed by two layers of BiLSTM. A 53-layer residual network is used to extract features, where each residual unit consists of a $ 1 \times 1$ convolution and a $ 3\times 3$ convolution operations. The first two residual blocks are down-sampled with $ 2 \times 2$ stride while the left three with $ 2 \times 1$ stride.

\subsubsection{Bidirectional Decoder}
The previous works on STR with a bidirectional decoder all treat the forward and backward parts separately. However, it is obvious that forward and backward processes are almost the same, which means nearly half of the decoder network parameters are redundant. So in this section, we propose a new method to decode the sequence context information bidirectionally with only one decoder.  

We need to capture the direction information for the unique decoder network, so the directional embedding is added apart from the one hot embedding.

After the GRU network iterate for T steps, a sequence $(y_1, y_2,\cdots, y_T)$ is predicted.
the context information $c_t$ can be encoded with the following attention mechanism:
\begin{equation}
\begin{split}
p_{t,i}&=\bm w^T\tanh(\bm W \bm s_{t-1} + \bm V \bm h_i + b)\\
\alpha_{t,i} &= e^{p_{t,i}} / \sum_{i'=1}^n{e^{p_{t,i'}}}\\
\bm c_t &= \sum_{i=1}^n \alpha_{t,i}\bm h_i
\end{split}
\end{equation}
where $s_{t-1}$ is the internal state.
After the attentional context is generated, we need to combine the embedded information. Different from ordinarily concatenating the context with one-hot embedding $f$, here we add one more direction embedding $d$ to distinguish forward and backward decoding process.
\begin{equation}
\begin{split}
(\bm x_t, \bm s_t) = \text{GRU} (\bm s_{t-1}, \text{cat}(&\bm c_t, f(y_{t-1})+\\&d(forward/backward)))
\end{split}
\end{equation}
where $x_t$ is used to predict $y_t \sim \text{softmax}(\bm W_o\bm x_t+b_o)$.

The loss function of the bidirectional decoder is:
\begin{equation}
Loss = \lambda l_{forward} + (1-\lambda) l_{backward}
\end{equation}
where $\lambda$ is a scaling factor used to define the importance between the forward and backward losses.

In the inference stage, we use beam-search with k candidates to increase the accuracy of decoding.

\subsection{Implementation Details}
The input image is resized to $36\times 128$ and the rectified image or the input of encoder is resized to $32\times 100$ pixels. Table \ref{ARN} is the detailed architecture of the sequence recognition network. $\lambda=0.7$ is configured for forward loss weight, the candidate of beam-search is $k=5$. the batch size is 64.

\begin{table}[htbp]
    \centering
    \begin{tabular}{llll}
        
        \hline
        & Layers  & Out Size & Configurations \\
        \hline
        \multirow{8}{*}{\rotatebox{90}{Encoder}}& Input  & $32 \times 100$ & k3, s1, 32     \\
        & Block1  & $16 \times 50$ & k3, s2, 64     \\
        & Block2  & $8 \times 25$ & k3, s2, 128     \\
        & Block3  & $4 \times 25$ & k3, s$2\times 1$, 256     \\
        & Block4  & $2 \times 25$ & k3, s$2\times 1$, 512     \\
        & Block5  & $1 \times 25$ & k3, s$2\times 1$, 512     \\
        
        & BiLSTM  & 25  & $256 \times 2$ hidden units \\
        & BiLSTM  & 25 & $256 \times 2$ hidden units \\
        \hline
        \multirow{4}{*}{\rotatebox{90}{Decoder}}
        & & &\\
        & GRU   & *  & \begin{tabular}[x]{@{}l@{}}256 attention units,\\256 hidden units \end{tabular}\\
        & & &\\
        \hline
    \end{tabular}
    \caption{Architecture of Recognition Network. Each of the block is a residual one, the attentional GRU decoder is a bidirectional decoding layer.}
    \label{ARN}
\end{table}

The Embedding dimension is 512, the target 68 characters covers 10 digits, lower case letters and 32 ASCII punctuations. The direction embedding is the same across all the 512 dimensions.

\section{Experiments}

In this section we show the various training and evaluation datasets and the extensive experiments results.

\begin{table*}[th]
    \centering
    \begin{tabular}{lccccccc}
        \hline
        Methods  & IIIT5k & SVT & IC03 & IC13  & IC15 & SVTP & CUTE \\
        \hline
        \cite{jaderberg2016reading} & - & 80.7 & 93.1 & 90.8 & - & - & - \\
        \cite{shi2016an} & 81.2 & 82.7 & 91.9 & 89.6 & - & - & - \\
        \cite{shi2016robust} & 81.9 & 81.9 & 90.1 & 88.6 & - & 71.8 & 59.2 \\
        \cite{lee2016recursive} & 78.4 & 80.7 & 88.7 & 90.0 & - & - & - \\
        FAN & 87.4 & 85.9 & 94.2&  \textbf{93.3} & 70.6& -& - \\
        ESIR  & 93.3 & \textbf{90.2} & - & 91.3    & 76.9 & 79.6 & 83.3    \\
        MORAN & 91.2 & 88.3 & \textbf{95.0} & 92.4  & 76.1 & 78.5 & 79.5 \\
        ASTER  & 93.4 & 89.5 & 94.5 & 91.8  & 77.8 & 79.7 & 81.9  \\
        \hline
        Firbarn & \textbf{93.4} & 89.8 & 93 & 92.7 & \textbf{81.3} & \textbf{80.8} & \textbf{86.1} \\
        \hline
    \end{tabular}
    \caption{Recognition results comparison  to state-of-the-art. Some methods use private dataset and achieve outperforming results, which are not fair and exclude in this comparison. the configure for rectification grid size is $3\times 10$ and the curve order is 4.}
    \label{comparision_rx}
\end{table*}

\subsection{Datasets}
We train the model on 2 general datasets, then evaluate on 7 benchmarks, which consist of 4 regular text datasets and 3 irregular text datasets, to show its rectification ability on curved, distorted and oriented text. A brief description of these datasets is as follows.
\subsubsection{Training Datasets}
Real-world labeling data is costly, following most scene text recognition works, we use the synthetic data as an alternative. Specifically, we train the model on the following two datasets, no more data is used.

\begin{itemize}
\item Synth90K \cite{jaderberg2014synthetic} contains about 7.2 million training images and it has been widely used for training scene text recognition models. After cropping with the lexicon, 9 million images are used for training.
\item SynthText \cite{gupta2016synthetic} is the synthetic image dataset that was created for scene text detection research. It has been widely used for scene text recognition research as well by cropping text image patches using the provided annotation boxes. 7 million cropped images are for training.
\end{itemize}

\subsubsection{Evaluation Datasets}

\begin{itemize}
\item IIIT5K-Words (IIIT5K) \cite{mishra2012scene} contains 3,000 images for testing, which are cropped from online scene texts images.
\item Street View Text (SVT) \cite{wang2011end} consists of 647 testing images, which are collected from the Google Street View. Many images are heavily distorted by noise or low-resolution.
\item ICDAR 2003 (IC03) \cite{lucas2003icdar} contains 860 images of cropped words after filtering. Following \cite{wang2011end}, words with non-alphanumeric characters or less than three are filtered.
\item ICDAR 2013 (IC13) \cite{karatzas2013icdar} inherits most of its data from IC03 and covers 1,015 cropped word images in total.
\item ICDAR 2015 (IC15) \cite{karatzas2015icdar} is collected via Google Glasses without careful positioning and focusing. 2,077 images with various corruptions in this dataset are used for testing.
\item SVT-Perspective (SVTP) \cite{quy2013recognizing} is specifically proposed to evaluate the performance of perspective text recognition algorithms. It consists of 645 images for testing.
\item CUTE80 (CUTE) \cite{risnumawan2014robust} is designed to evaluate curved text recognition. 288 cropped images are used for testing.
\end{itemize}

\subsection{Evaluation Metrics}
Following the protocol and evaluation metrics that have been widely used in scene text recognition research \cite{shi2018aster,baek2019what}, we evaluate the model performance using the case-insensitive word accuracy. The evaluation only counts the letters and digits error, no lexicons are used.

\begin{figure}[htbp]
    \centering
    \includegraphics[width=.4\textwidth]{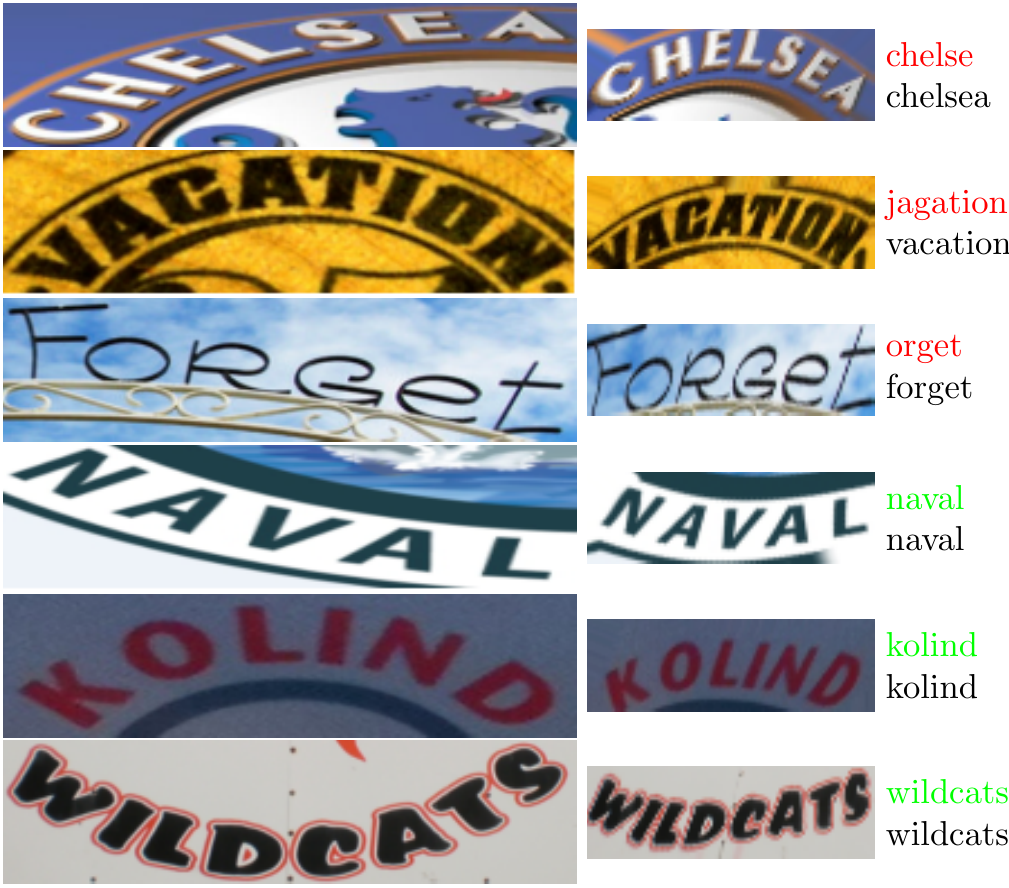}
    \caption{Examples of the evaluation on CUTE. Left side are orignal images and the middle are rectified one. The black color denote the ground truth, red is wrong, green is right.}
    \label{examples_c}
\end{figure}

\subsection{Comparisons with Rectification Methods}

Table \ref{comparision_rx} is the full results of the comparison with previous state-of-the-art works, which shows our method is quite effective on the irregular datasets and outperform the previous results.

\begin{table}[htbp]
    \centering
    \begin{tabular}{llll}
        \hline
        Order  & IC15 & SVTP & CUTE \\
        \hline
        2  & 79.2 & 80 & 80.2 \\
        3 & 79.3 & 80.6 & 80.5 \\
        4 & 79.5 & 80.8 & 81.2 \\
        5 & 79.9 & 80.8 & 82.6 \\
        \hline
    \end{tabular}
    \caption{Compare for the curve orders.}
    \label{tab:order}
\end{table}

\begin{table}[htbp]
    \centering
    \begin{tabular}{llll}
        \hline
        size  & IC15 & SVTP & CUTE \\
        \hline
        $2\times 18$  & 80.6 & 80.8 & 84.6 \\
        $3\times 12$ & 81.2 & 80.6 & 85.2 \\
        $4\times 9$ & 80 & 79.6 & 83.3 \\
        \hline
    \end{tabular}
    \caption{Compare for the grid size.}
    \label{tab:size}
\end{table}

\section{Conclusion}

A finer rectification network before the general STR helps to rectify warped images. The simplified one-layer bidirectional decoder is shown to be effective with the separated two decoders. Extensive evaluations show this end-to-end model is more effective, especially on irregular scene text images. Experiments on the grid size and the order of curve show our method can improve the recognition accuracy of warped images greatly. The further work can combine more image information to predict the text boundary and train with highly-curved text.


\bibliographystyle{named}
\bibliography{firbarn}

\end{document}